\documentclass[journal]{IEEEtran}
\usepackage{graphicx}
\usepackage{cite}
\usepackage{picinpar}
\usepackage{amsmath}
\usepackage{url}
\usepackage{flushend}
\usepackage[latin1]{inputenc}
\usepackage{colortbl}
\usepackage{soul}
\usepackage{multirow}
\usepackage{pifont}
\usepackage{color}
\usepackage{alltt}
\usepackage[hidelinks]{hyperref}
\usepackage{enumerate}
\usepackage{siunitx}
\usepackage{breakurl}
\usepackage{epstopdf}
\usepackage{pbox}
\usepackage{algorithm}
\usepackage{algorithmicx}
\usepackage[noend]{algpseudocode}
\usepackage{array, multirow, bigdelim, makecell, booktabs} 
\begin{document}
\title{Label-Attention Transformer with Geometrically Coherent Objects for Image Captioning }

\author{
	Shikha Dubey\textsuperscript{1}, 
	Farrukh Olimov\textsuperscript{2}, \\
	Muhammad Aasim Rafique\textsuperscript{1},
	Joonmo Kim\textsuperscript{3},
	\\ and Moongu Jeon \IEEEauthorrefmark{2}\textsuperscript{1}, 
    \thanks{
        \textsuperscript{1} School of Electrical Engineering and Computer Science, Gwangju Institute of Science and Technology (GIST), Gwangju, 61005, South Korea;
      }
      \thanks{ \textsuperscript{2} Threat Intelligence Team, Monitorapp, Seoul, South Korea;}
      \thanks{
      \textsuperscript{3} Dankook University, Department of Computer Engineering, Jukjeon, South Korea;}
  \thanks{\IEEEauthorrefmark{2} represents the corresponding author}
}	
\maketitle

\begin{abstract}
Automatic transcription of scene understanding in images and videos is a step towards artificial general intelligence. Image captioning is a nomenclature for describing meaningful information in an image using computer vision techniques. Automated image captioning techniques utilize encoder and decoder architecture, where the encoder extracts features from an image and the decoder generates a transcript. In this work, we investigate two unexplored ideas for image captioning using transformers: First, we demonstrate the enforcement of using objects' relevance in the surrounding environment. Second, learning an explicit association between labels and language constructs. We propose label-attention Transformer with geometrically coherent objects (LATGeO). The proposed technique acquires a proposal of geometrically coherent objects using a deep neural network (DNN) and generates captions by investigating their relationships using a label-attention module. Object coherence is defined using the localized ratio of the geometrical properties of the proposals. The label-attention module associates the extracted objects classes to the available dictionary using self-attention layers. The experimentation results show that objects' relevance in surroundings and binding of their visual feature with their geometrically localized ratios combined with its associated labels help in defining meaningful captions. The proposed framework is tested on the MSCOCO dataset, and a thorough evaluation resulting in overall better quantitative scores pronounces its superiority.  
\end{abstract}


\markboth{}%
{}
\definecolor{limegreen}{rgb}{0.2, 0.8, 0.2}
\definecolor{forestgreen}{rgb}{0.13, 0.55, 0.13}
\definecolor{greenhtml}{rgb}{0.0, 0.5, 0.0}

\section{Introduction}
\label{sec:introduction}

\IEEEPARstart{I}{mage captioning} is one of the core problems in scene understanding and it leverages the progress in computer vision (CV) and natural language processing (NLP). It manifests the inherent challenges of spatial, temporal, and sequential data modalities.  Another obtrusive challenge is the translation from a spatial modality to a sequential modality that arborizes likely transcriptions of a scene. A widely adopted solution is to use an encoder-decoder architecture where an encoder extracts features, and a decoder transcribes the captions. Moreover, in the deep learning era, convolution neural networks (CNNs) and recurrent neural networks (RNNs) are adopted for encoding and decoding, respectively. RNNs are frequently used with attention layers to preserve and capture the distant association in sequential data \cite{A55,A6}. An inherent bound in a language is the length of sentences which does not allow the use of an attention layer with feed-forward networks and needs sequential modeling of them. However, SOTA algorithms showed promising results, they do not learn relationships among objects and surroundings and the class to label association.
\begin{figure*}
\begin{center}
\includegraphics[trim={0.0cm 8.0cm 17.0cm 0.0cm}, clip=true, height=8.0cm, width=14.5cm]{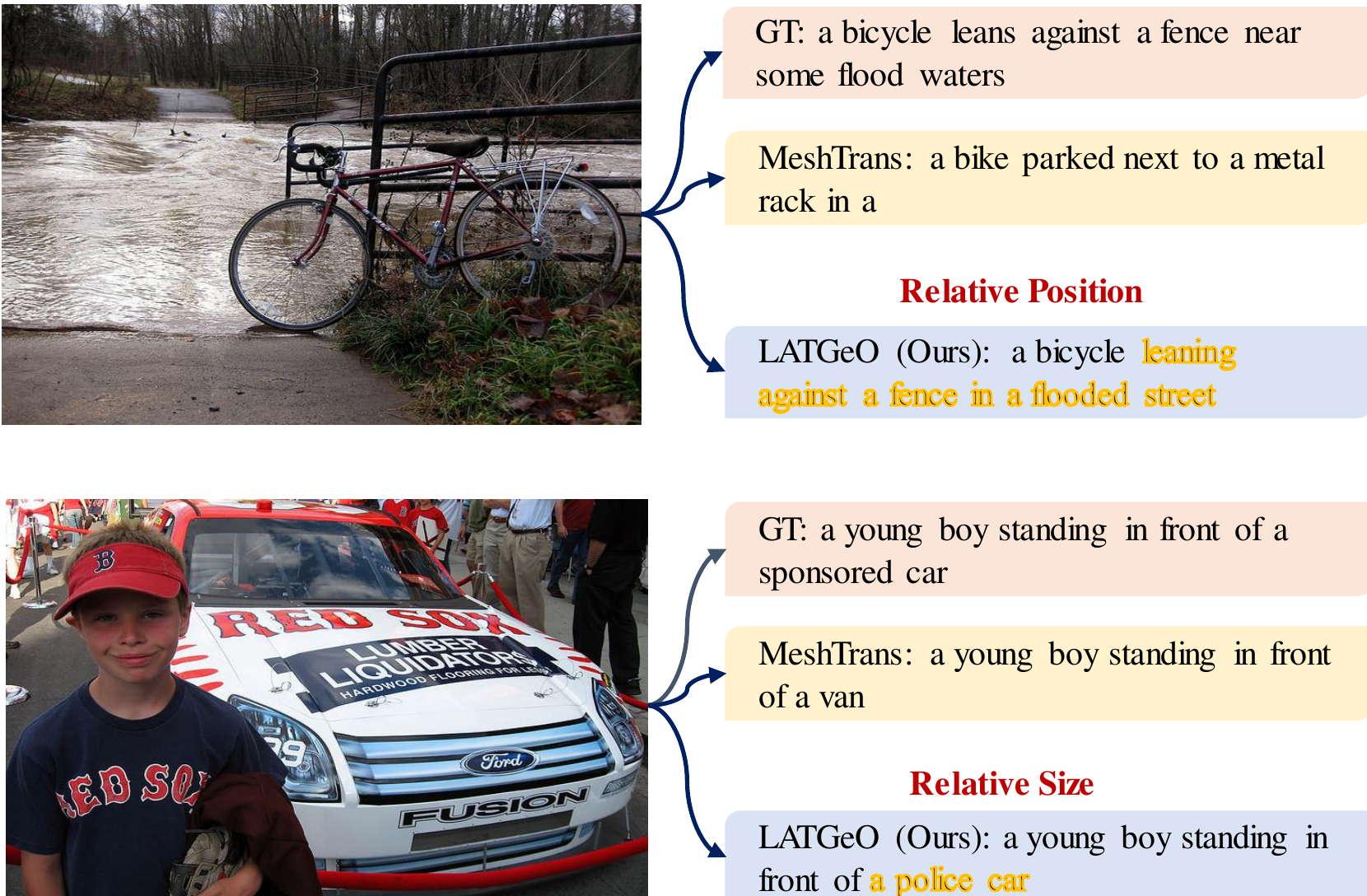}
\end{center}
\caption{Examples of image captioning: signifying utility of background information and geometrically localized ratio.}
\label{intromod}
\end{figure*}

This study proposes a novel technique that uses concrete features of objects and their surroundings, the localized ratio of objects using their geometrical properties, and uses their compliance with the language rules with a label-attention module for image captioning. In particular, we assimilate high-level cognition by generating proposals from images, utilizing the available geometrical formations of proposals, and learning their relationship with surroundings and labels. The proposals are common identifiable vision interpretations and are detected objects in images. The learning of the proposed system is inspired by the recent advancements in encoder-decoder neural networks with self-attention layers called transformers. Variants of transformers are in active use in image captioning and are discussed in detail in section \ref{RelW}. Briefly, this study proposes a novel architecture that uses a label-attention transformer with geometrically coherent objects (LATGeO).

LATGeO expounds on using geometrically coherent object proposals and label-attention to learn the relationship among objects. The object detector is used for extracting object proposals. The main idea of extracting objects from images is to provide the proposed architecture with fine-grained information about the content of the image along with the entire image, while the geometrical properties identify the association among objects. The geometrically coherent properties are encapsulated for better learning of the relative positions and size of objects; for example, to learn the relative size of objects from ``a young boy standing in front of a van" to ``a young boy standing in front of a car" and to learn the relative positions of objects from ``a bike parked next to a metal rack in a" to ``a bicycle leaning against fence in a flooded street" (consider Fig.~\ref{intromod}). A similar study is considered in \cite{A2}, but our method considers the ratio of objects' dimensions. The visual clues are accumulated and processed with label attention to comprehend the language semantics with a decoder module in LATGeO and generates meaningful image captions. In a transformer composition, multiple encoders are stacked, and the output of an encoder is passed to the next encoder in the stack. Usually, the output of the last encoder in a stack is passed to the first decoder in similarly stacked decoders. However, a recent study \cite{A20} discusses a composition of an encoder stack fully connected to a decoder stack. The fully connected composition inspires LATGeO as it explores multi-level geometrical and visual representations of objects in an image.

Contributions of this study are summed up as follows:\begin{enumerate} 
\item We propose a transformer-based framework LATGeO for image captioning task, which encapsulates multi-level visual and geometrically coherent proposals to establish the relationship among objects based on their localized ratios.\footnote{The code will be publicly available on \url{https://github.com/shikha-gist/Image-Captioning/}.}
\item LATGeO uses object proposals and relates its embeddings with less significant surroundings to discover object coherence. 
\item A novel label-attention module (LAM) is proposed, which is an extension of the standard transformer to bridge the gap between visual and language domains. In LAM, object labels are associated with each decoder layer's input as prior information for caption generation.
\item Our extensive experiments on the MS COCO dataset show enhanced results compared to other attention models by simply using a single model.
\item An ablation study includes a comprehensive study of using different object proposal methods and compositions of encoder-decoder layers and the impact of individual proposed modules of LATGeO. 
\end{enumerate} 
This paper is composed as follows: Section \ref{RelW} details the recent developments in image captioning research, and section \ref{method} exposits the proposed architecture LATGeO. Section \ref{impleval} details extensive experiments performed to support the proposed methodology with ablation studies, and section \ref{con} briefs the conclusion of this work.

\section{Related Works} \label{RelW}
Machine vision evolved over the past two decades and extends to resolve challenging problems in scene understanding like image captioning problems. Image captioning fuses the progress in cutting-edge approaches from CV and NLP. The progress in image captioning techniques is categorized  into four sections in our review work: template-based techniques, deep neural network-based image captioning, attention-based techniques, and transformer-based image captioning techniques.
\subsection{Template-based image captioning} 
Conventional algorithms are based on two approaches: In earlier approaches like in \cite{A68}, image retrieval techniques retrieve images using a few collections of keywords as templates from image-caption data pairs and generate captions for the retrieved images using annotated captions. In comparison, later approaches \cite{A67} practice bottom-up algorithms that infer sentence parts like nouns, verbs, and adjectives from images and apply pre-defined caption templates to generate the image descriptions. Recently, Lu et al. \cite{A18} have adopted specific image regions explicitly bound to slot locations of template captions, and the template slots are packed with visual features of the extracted objects. Template-based captioning methods require human-crafted templates, which limits the generalization of these techniques. 

\subsection{Deep neural network-based image captioning} 
Since the beginning of the recent AI spring a decade ago, deep neural networks have influenced all branches of AI, and image captioning is not an exception \cite{A17,A14,A59,A11,A36}. Earlier CNN and RNN (LSTM) in an encoder-decoder composition created a remarkable impact. Early works such as \cite{A10,A11}, image captioning are treated as a conventional machine translation problem by transforming an image into N-dimensional vector representation and feeding it as input into the RNN decoder. Vinyals et al. \cite{A11} applies a deep CNN network to encode vision features from the whole image and utilizes RNN to generate captions by maximizing the likelihood of target caption. A constraint of this procedure is that it is challenging to represent all objects present in an image and their attributes as a single feature vector; consequently, scene graphs and object detection techniques are incorporated in image captioning to address this constraint \cite{A8,A36,A14}. Likewise, the attributes information is additionally added to the RNN input to learn the relationship among objects \cite{A14,A36}.   

Model optimization plays a vital role in training DNN; recently, \cite{A13,A5,A58} have suggested improvements in the optimization technique for training. \cite{A13} attempts to boost the training using actor-critic reinforcement learning to optimize non-differentiable quality metrics. Our method also incorporates a similar boosting training technique to advance performance. Lately, policy gradient techniques \cite{A59,A5} for reinforcement learning in such tasks also exhibited improved performance. Furthermore, Liu et al. \cite{A5} propose a Context-Aware Visual Policy network caption generation employing an actor-critic policy gradient method and visual attention. Similarly, \cite{A58} Ren et al. introduce a policy network comprising a convolutional and recurrent neural network and a value network consisting of CNN, RNN, and MLP to generate captions.

\begin{figure*}
\begin{center}
\includegraphics[trim={0.0cm 6.0cm 0.0cm 0.0cm}, clip=true, height=5.0cm, width=14.2cm]{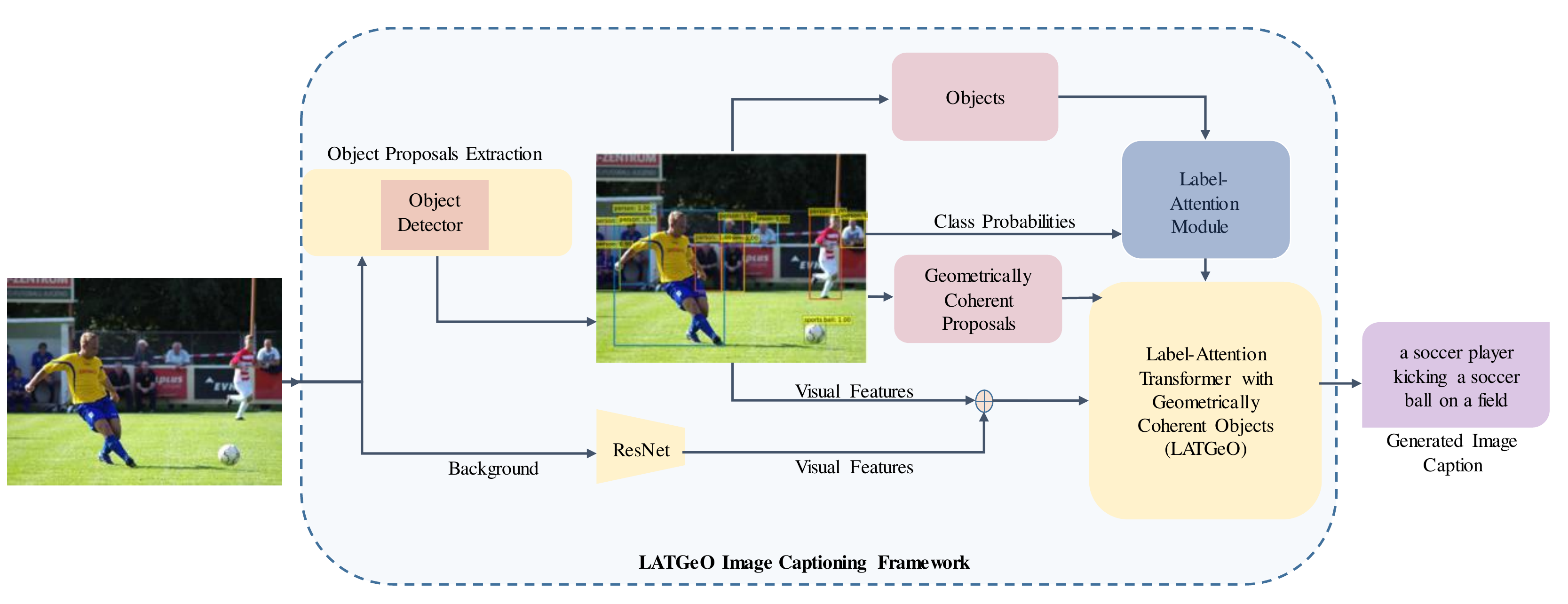}
\end{center}
\caption{The proposed architecture for image captioning, LATGeO.}
\label{model}
\end{figure*}
\subsection{Attention-based image captioning}
The attention layers added to the DNN particularly advance the results of sequential learning tasks and invigorate recent image captioning developments \cite{A57,A3,A51,A52,A44,A12,A34,A40}. Earlier, Xu et al. \cite{A51} have introduced a spatial attention model using image feature maps to generate image captions extended with a channel-wise attention module in \cite{A52}. Later, \cite{A44} introduces a gated hierarchical attention module by merging low-level features with high-level features. Xu et al. \cite{A33} propose an attention-gated LSTM model where the output gate incorporates visual attention and forwards to the cell state of LSTM. Lu et al. \cite{A12} introduce an adaptive attention model on visual sentinel by deciding which region of an image should be attended for extracting meaningful image features to generate sequential caption words. To learn the multi-level dependencies in objects, \cite{A47} proposes a multi-stage image captioning model consists of one convolutional encoder and multiple stacked attention-based decoders to generate fine captions. In our proposed technique, multi-level dependencies are learned by a single transformer. A series of recent works address the relational reasoning among regions using various compositions of activation layers in RNN and CNN \cite{A34,A7,A60,A53,A54,A55,A56}. Yao et al. \cite{A9} have introduced a hierarchical parsing of detection and segmentation of objects into a tree structure and used it as input to the encoder. Huang et al.\cite{A45} propose the refinement network to correlate semantic information with attributes to improve image captions. Our proposed algorithm has learned such a relationship using geometrical information of the extracted objects without additional attributes similar to \cite{A2}.

To utilize the individual object's features for more reliable context learning, object detection and attention module are combined in \cite{A51,A4,A46}. These studies encode the visual features of extracted objects and transfer them to the recurrent network with attention for generating captions of the image. Whereas \cite{A4} also has additional information of object attributes to refine the predicted captions. \cite{A46} utilizes a convolutional graphical model to represent structured information in the form of detected objects and their relationships. These features are passed through a hierarchical attention-based module for caption generation at each time step. \cite{A49} also studies Hierarchical-Attention by using GAN  based model. Recent algorithms \cite{A6} propose the visual relationship attention on extracted objects region and investigate the visual relationship among them for generating captions. However, the algorithm \cite{A6} employs a Graph Convolutional Networks on detected objects and a Long Short-Term Memory network to generate captions based on the attention module. For model optimization, Rennie et al. \cite{A3} propose self-critical learning, which optimizes models based on evaluation metrics such as CIDEr, resulting in significant performance improvements over the methods that use cross-entropy objectives only.  

However, these attention mechanisms have exhibited promising results on the image captioning task but lack in learning the relationship among background and objects.    
\subsection{Transformer-based image captioning}
Transformers are state-of-the-art in NLP \cite{A1}, and image captioning adopted transformers for caption transcription. \cite{A31} has recently explored the transformer by learning attention module using contextualized embedding for individual regions and examined visual relationships by spatial object regions. \cite{A19} proposed additional attention on the multi-head attention of transformer for image captioning. A recent study \cite{A2} transfers encoded visual features of the extracted objects through the transformer architecture and for learning relative appearance among objects. It also passes extracted encoded information of objects' bounding boxes along with the visual information through a multi-head self-attention mechanism. In comparison, \cite{A20} learns such dependencies among the regions of interest with the help of memory-augmented attention. Our proposed method, LATGeO, is motivated by these recent works \cite{A2,A20,A15} of self-attention mechanisms though diverse in many aspects. LATGeO learns the relationship among objects as well as a relationship with a background unlike any previously proposed algorithms and transformer-based algorithms \cite{A2,A20}. 

Furthermore, LATGeO proposes different geometrically coherent features using localized ratios which is distinctive from the method proposed in \cite{A2} and utilizes the label-attention modules in the decoder of the multi-head attention module additional to meshed transformer proposed in \cite{A20}. Further, we have experimented with LATGeO using several object proposal methods like DETR \cite{A15}, which is a newly proposed transformer-based object detection method. The proposed technique learns relationships among objects and their context without any additional information like attributes or semantics. Aspects of our proposed approach are presented in the following sections.

\section{Method} \label{method}
This work proposes a label-attention transformer which uses geometrically coherent objects for image captioning. The complete framework is depicted in Fig.~\ref{model}. LATGeO explores tangible objects' features in an image at a multi-level fine-grained representation of object features to generate meaningful transcriptions. Firstly, we extract objects from an image that are called proposals in this study. Secondly, the proposals are assigned to labels from the known classes, and the labels are passed through a label-attention module. Thirdly, an effective geometrical relationship of the detected proposals is computed. A multi-level representation of the objects and the less significant features are passed as input to the final learnable block of $($LATGeO$)$, a fully connected encoder-decoder transformer. The decoder generates an image caption in the end. The details of each component of the proposed framework are explained in subsequent subsections. 
\begin{figure*}
\begin{center}
		\includegraphics[height=8.5cm, width=13.0cm]{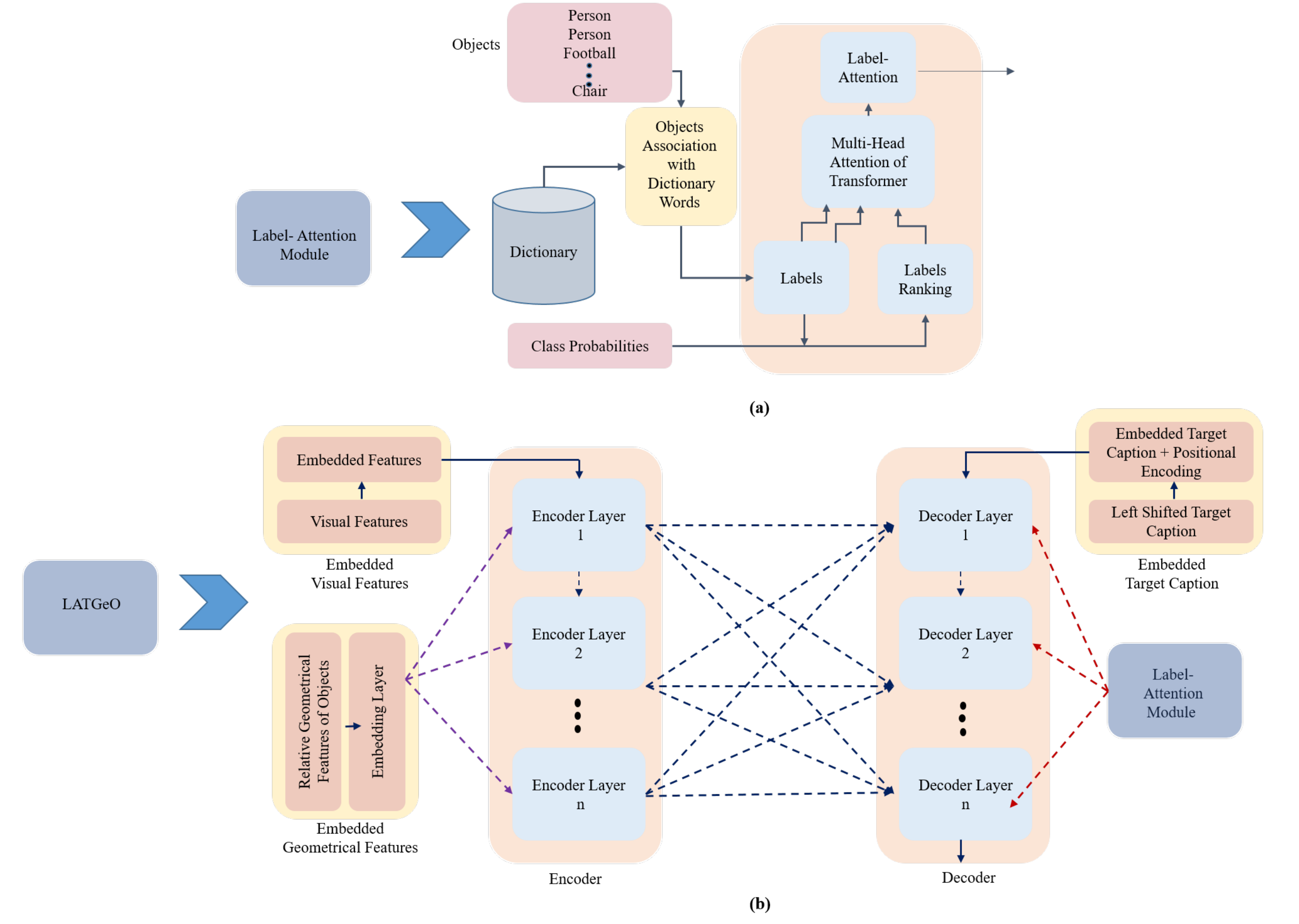}
\end{center}
\caption{$($a$)$ Configuration of label-attention module (LAM) $($b$)$ Configuration of LATGeO.}
\label{model_2}
\end{figure*} 

\subsection{Object Proposals and Background} \label{op}
Proposals are essential components of the core strategy presented in this study, and numerous choices of a proposal with distinctive features are deliberated. A valuable proposal is translated into invariant and covariant features, and a semantic relationship is defined among proposals. It is empirically decided to use object detection in images to generate meaningful proposals. The proposals are generated from the detected objects as $2048-$dimensional visual features. Furthermore, each detected object's class probabilities, label(s), and geometrical features are explored for their use in image captioning. The labels generated from object proposals are fed to the label-attention module (LAM), and geometrical features are fed to the transformer module in LATGeO.

This study has generated proposals data with the SOTA DNN object detectors, i.e., Faster R-CNN \cite{A22}, and DETR \cite{A15}. Faster R-CNN is a two-stage object detection model, a base CNN model, ResNet in our study, for features extraction in the first stage, and a region proposal network $($RPN$)$ is utilized to generate the bounding boxes using the intersection over union $($IoU$)$ method in the second stage.

The second DNN network for proposal generation employed in this study is DETR, which uses a transformer architecture for multiple objects detection. Faster R-CNN is potentially fine-tuned for image captioning domain-specific, making it suitable for further use in our detailed experiments. Performance comparison of our proposed architecture, LATGeO, using both object detectors, DETR and Faster R-CNN, is given in ablation study section \ref{DETR} to assert the choice later.

In addition to the proposals, visibly less significant features are also fed to LATGeO, which render the relationships between objects and background. A pre-trained ResNet \cite{A23} model is used to extract the background features, and the features are progressed through the LATGeO block.

\subsection{Geometrically Coherent Proposals } \label{gf}
Image captions are hard to generate with proposals features and labels only; therefore, this study adopts a natural coherence among proposals. The coherent relationship is computed using the geometrical properties of the bounding boxes of the proposals detected by Fast R-CNN in the geometrically coherent proposals (GCP) block. A pair $($a$,$b$)$ of the detected objects are fed to a GCP block where the relative geometrical coherence $\xi \left(a,b\right)$ of the proposals is calculated using equation $($\ref{eq1}$)$.  
\begin{multline}
\label{eq1}
\hspace*{0.4cm}\xi \left(a,b\right)= \left( log \left(\frac{x_a}{x_b}\right), log \left(\frac{y_a}{y_b}\right), \right. \\
\left. log \left(\frac{w_a}{w_b}\right), log \left(\frac{h_a}{h_b}\right) \right), 
\end{multline}
where $(x_a, y_a), (x_b, y_b)$ are center coordinates, $(w_a, h_a)$ and $(w_b, h_b)$ are widths and heights of objects $a$ and $b$, respectively. Semantically, equation $($\ref{eq1}$)$ gives a simple ratio of the scale dimensions $($Ratio-Comparison$)$ of two objects, different from $($L1-Comparison$)$ of objects used in \cite{A2}. 
The weights for attention mechanism utilizing these geometrical features for further processing are calculated below:  
\begin{equation}\label{eq3}
   \hspace*{-3.0cm} \eta_G^{ab}=ReLU(Emb(\xi)w_G),
\end{equation}
where $Emb()$ in equation $($\ref{eq3}$)$ represents embedding of objects' relative geometrical features. $Emb()$ is a learned embedding that projects objects' relation vector $(\xi \left(a,b\right))$ into a high-dimensional embedding, similar to the study \cite{A2}. $w_G$ is a learned $d_{model}-$dimensional vector that projects these vectors down to a scalar. These geometrical features are then propagated through the LATGeO block. 

\subsection{Label-Attention Module} \label{la}
It is challenging to transcribe the features extracted from images into meaningful captions because they have more transcriptions than their equivalent detected proposals. Therefore, meaningful labels from language models are considered in this study to reduce the combinations of transcription of the extracted features by passing it through a label-attention module $($LAM$)$. LAM learns the association of labels and the detected proposals and attends the meaningfully related classes of proposals. LAM generates embedding of labels and classes, which are fed to the LATGeO module. The detailed working of LAM is as follows:

First, LAM associates all the detected objects' classes with the available dictionary $D$ and generates labels using the following equation:
\begin{equation}
    \hspace*{-1.0cm}L\_O^{i} = emb(Index(w^j:C\_O^{i} == D(w^j))),
\end{equation}\label{eq16}
where $L\_O^{i}$, and $C\_O^{i}$ represent the label, and a class of the $i^{th}$ object, respectively. $D(w^j)$ represents the $j^{th}$ word present in dictionary $D$. $emb()$ converts label of the $i^{th}$ object into high-dimensional embedding of dimension $d\_model$. Afterward, LAM adjusts the rank of associated labels using the class probabilities of detected proposals using the following equation:
\begin{equation}\label{eq17}
  \hspace*{-3.7cm}R\_O^{i} = L\_O^{i} * Pr(C\_O^i),
\end{equation}
where $R\_O^{i}$ represents the ranking of the $i^{th}$ object among all detected objects and $Pr(C\_O^i)$ represents the probability of the $i^{th}$ object's class. Next, we pass all these embedded labels and their ranking as queries $(Q)$ and keys $(K)$ respectively to the multi-head attention module of the vanilla transformer \cite{A1} as given below:
\begin{equation}\label{eq18}
    \hspace*{-0.5cm}L\_Att =  \sigma(Multi\textendash Head(L\_O, R\_O,L\_O)),
\end{equation}
where equation (\ref{eq18}) represents the label-attention by taking sigmoid $(\sigma())$ of Multi$\textendash$Head attention output.
\begin{multline}
\label{eq19}
 \hspace*{0.4cm}Multi\textendash Head(Q,K,V)= Concat(Head\_1,\\ 
 Head\_2,..Head\_h)W^0,
\end{multline}
where equation (\ref{eq19}) concatenates the output of all the attention heads, and $W^0$ is the learned projection matrix for multi-head attention. $V$ stands for value and $Head_h$ can be calculated as follows:
\begin{equation}\label{eq20}
  \hspace*{-2.0cm}Head_h = Att(QW_{j}^{Q}, KW_{j}^{K}, VW_{j}^{V}),
\end{equation}
where $W_{j}^{Q}$, $W_{j}^{K}$, and $W_{j}^{V}$ are the projection matrices for queries $(Q)$, keys $(K)$, and values $(V)$, respectively. $Att(:)$ is described later in equation (\ref{eq9}).
Later this label-attention $(L\_Att)$ is joined with the output of each encoder layer $(E\_Out(n))$ as given in equation (\ref{eq21}) and passed as an input to the decoder layer $(D\_Inp)$ of LATGeO.
\begin{equation}\label{eq21}
  \hspace*{-0.7cm}D\_Inp\_L\_Att = {GS}_{n=1}^{L}(E\_Out(n)* L\_Att),
\end{equation}
where $L$ is the number of encoder layers, and $GS$ is the sigmoid gating, similar to the study \cite{A20}.
\begin{figure*}
\begin{center}
\includegraphics[trim={5.5cm 4.5cm 8.0cm 0.0cm}, clip=true, height=5.5cm, width=8.5cm]{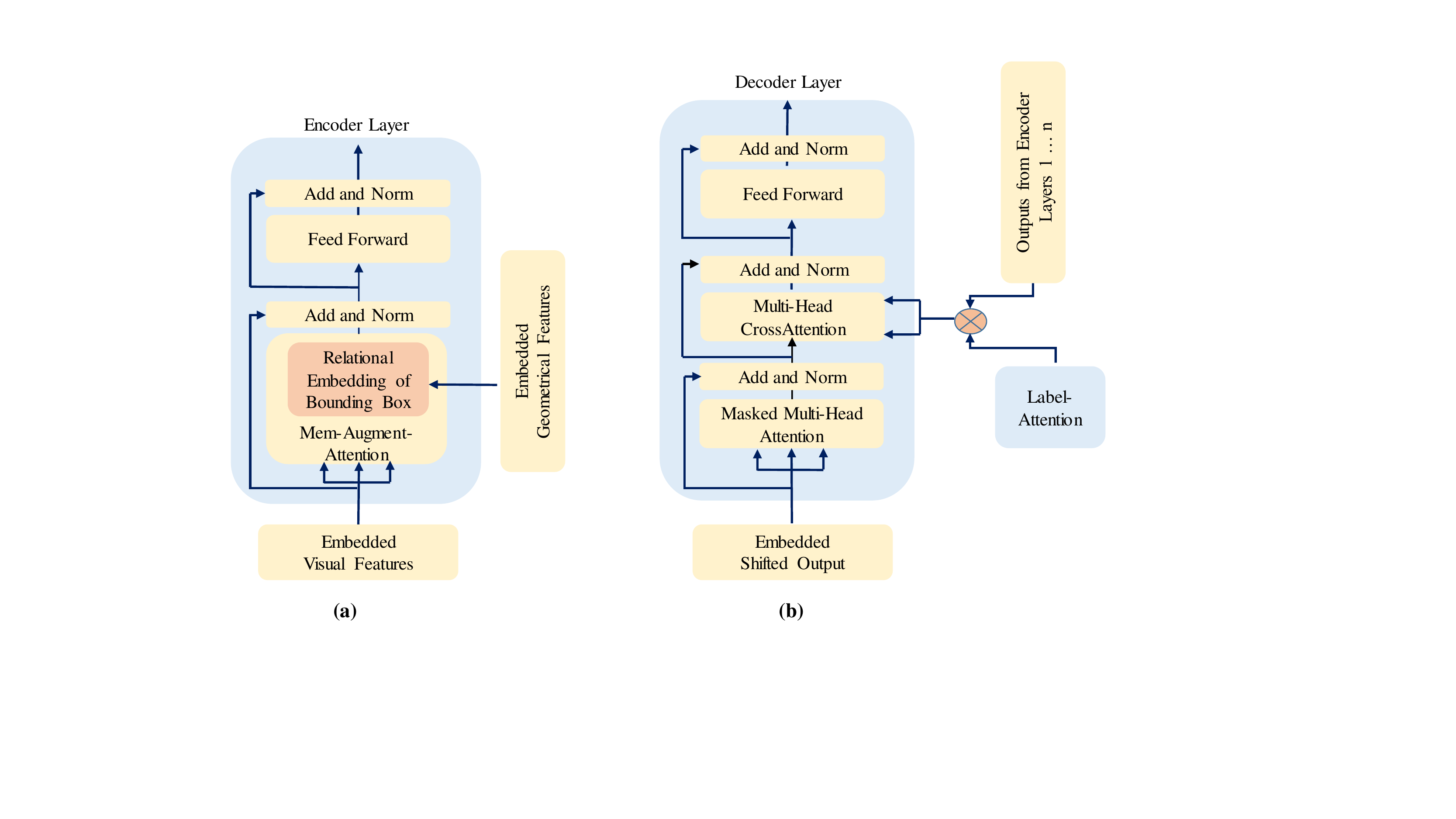}
\caption{$($a$)$ A detailed structure of the encoder layer. $($b$)$ A detailed structure of the decoder layer.}
\label{oit}
\end{center}
\end{figure*}
\subsection{LATGeO} \label{oirtb}
Detailed configuration of LATGeO block is presented in Fig.~\ref{model_2} $($b$)$ and Fig.~\ref{oit}. This block of transformer takes embedding of proposals, background features, LAM features, and GCP features as its input and generates caption of the image as output. The embeddings of proposals, background, and GCP are the encoder inputs, and the LAM embedding is an input of the decoder. The details are provided in the following encoder and decoder sections.  

\subsubsection{Embedded Visual Features}
Features of objects' proposals and background are concatenated together $($shown in Fig.~\ref{model}$)$ and then projected down to a $d_{model}$-dimensional vector as an embedding using trainable matrix $W$. This generates embeddings of the visual features $($shown in Fig.~\ref{model_2} $($b$))$. 

\subsubsection{Encoder Layer} \label{el}
LATGeO's encoder is composed of $L$ identical encoder layers. Each layer is composed of two components, a multi-head memory-augmented-attention $(MA\_Att)$ and a position-wise feed-forward network, along with the residual connections between these components \cite{A20}. The detailed structure of the encoder layers of the LATGeO is shown in Fig.~\ref{oit} (a). The GCP embedding, along with the object proposals and background embedding, is utilized in a memory-augmented-attention $(MA\_Att)$ mechanism as follows:
\begin{equation}\label{eq25}
    MA\_Att(Q,K,V,G)=head(Q,K,V,\eta) 
\end{equation}
where $G$ represents GCP embeddings as $\eta$ and $head()$ can be calculated as follows:
\begin{equation}\label{eq6}
    \hspace*{-1.0cm} head(Q,K,V,\eta)=softmax\left(\eta \right)V,
\end{equation}
\begin{equation}\label{eq4}
    \hspace*{-2.3cm} \eta^{ab}=\frac{\eta_G^{ab}\exp(\eta_A^{ab})}{\sum_{l=1}^N \eta_G^{al}\exp(\eta_A^{al})},
\end{equation}
where $\eta_G$ is given in equation (\ref{eq3}) and $\eta_A$ can be calculated as follows: 
\begin{equation}\label{eq5}
    \hspace*{-4.0cm}\eta_A=\frac{QK^T}{\sqrt{d_k}},
\end{equation}

\begin{equation*}
    \hspace*{-4.0cm}Q = [W_qQ], 
\end{equation*}
\begin{equation*}
    \hspace*{-3.5cm}K = [W_kK,M_k], 
\end{equation*}
\begin{equation}\label{eq26}
    \hspace*{-3.7cm}V = [W_vV,M_v], 
\end{equation}
\begin{multline}
\label{eq27}
  \hspace*{1.0cm}Multi\textendash Head(Q,K,V,\eta)= Concat(head_1,\\ head_2,..head_h)W^0,
\end{multline}
In equation (\ref{eq4}), $\eta^{ab}$ represents the combined attention weights calculated by incorporating geometric attention weights from equation (\ref{eq3}) into the $MA\_Att$ mechanism. Also, $K, V$ represent key and value along with different trainable memory slots $M_k$ and $M_v$ of size $M$, respectively, as defined in equation (\ref{eq26}) like in \cite{A20}. $Q$ represents a key. Moreover, all the encoder layers are stacked so that the $l^{th}$ layer takes input from the previous layer ${(l\textendash1)}^{th}$, and each layer consists of $h$ number of $MA\_Att$ heads as given in equation (\ref{eq27}).   

\subsubsection{Decoder Layer}\label{dl}
LATGeO's decoder is composed of $L$ identical decoder layers. Each layer is composed of three components, a multi-head cross-attention, masked multi-head attention, and a position-wise feed-forward network, along with the residual connections among these components. The detailed structure of the encoder layers of the LATGeO is shown in Fig.~\ref{oit} (b). The decoder takes a sequence of previously generated words and multi-level representations from the output of all encoder layers and LAM embedding as input (equation (\ref{eq21})) and generates the next word. The attention mechanism in a decoder layer is similar to the study in \cite{A20} given as follows: 
\begin{equation}\label{eq7}
  \hspace*{-0.8cm}MA\_Att(\hat{I},Y)=\sum_i^N \alpha_i \times CrossAtt(\hat{I}^i,Y),
\end{equation}
\begin{equation}\label{eq8}
   \hspace*{-2.4cm} CrossAtt(\hat{I}^i,Y)=Att(Y,\hat{I}^i,\hat{I}^i),
\end{equation}
\begin{equation}\label{eq9}
   \hspace*{-1.9cm}Att(K,V,Q)=softmax\left(\frac{QK^T}{\sqrt{d_K}}\right)V ,
\end{equation}
\begin{equation}\label{eq10}
  \hspace*{-1.7cm}\alpha_i=\sigma(W_i\cdot[Y,CrossAtt(\hat{I}^i,Y)]+b_i)
\end{equation}
where equation (\ref{eq7}) represents the attention of a decoder layer, which is a weighted sum over cross-attention, $CrossATT$. The $CrossATT$ in equation (\ref{eq8}) is responsible for taking attention on the decoder layer's input $Y$ and all outputs of encoder layers after applying label-attention given in equation (\ref{eq21}) as $\hat{I}^i,\ i \in [1,2,...,L]$ and $i$ represents $i^{th}$ encoder layer. $\sigma$ represents the sigmoid activation function. Equation (\ref{eq9}) represents the attention module. Equation (\ref{eq10}) represents the calculation of the weights matrix $\alpha_i$, which computes the relevance between CrossATT and decoder layer input $Y$. Moreover, the current predicted word from the decoder layers depends only on previously predicted words; therefore, we apply a masked self-attention operation on the input target sequence similar to work \cite{A20}. Each decoder layer consists of $h$ number of attention heads.

\subsubsection{Encoder Decoder Connection}
The connections from encoder layers to decoder layers in LATGeO are shown in Fig.~\ref{model_2} $($b$)$, a fully connected encoder-decoder. The output of all encoder layers is concatenated using label-attention as given in equation (\ref{eq21}) and passed as input to every decoder layer after applying a sigmoid gating technique similar to the method in \cite{A20}. In this study, various compositions of the encoder-decoder connections are explored, and \ref{AddE} demonstrates single-connection, skipped-connections, and residual-connections compositions. The detailed structure of the encoder and the decoder layers of the LATGeO is shown in Fig.~\ref{oit}. A benefit of using the selected composition of the encoder-decoder in LATGeO is to attend the output of all encoder layers and reevaluate if the stack of encoders misses a valuable relationship. The effectiveness of the selected composition is further discussed with a demonstration of results in the \nameref{eval} section.
\subsection{Training Objective Functions} \label{ts}
LATGeO is trained using masked cross-entropy objective function $($XE$)$ similar to the studies \cite{A2} with an additional smoothing function. Afterward, it is tuned with reinforcement Learning $($RL$)$. The XE objective function, $L(\phi)$ is the sum of the negative log-likelihood of the correctly predicted words at each step given as follows:
\begin{equation}\label{eq11}
   L(\phi )= \\
   -\sum_{n=1}^{N} log\left( p_{\phi }(Sw_{n}|I,Sw_{1},...,Sw_{n-1}) \right)  
\end{equation}
where $\phi$ is the learning parameters, $Sw_n$ represents a one-hot vector for the $n_{th}$ word in a ground truth sentence or Image-caption of length $N$, and $I$ is an input image. The architecture is optimized on the best validation score of Cider-D metrics obtained after supervised learning. The reward function $r(.)$ of RL based on the CIDEr score of a generated caption is given in equation (\ref{eq12}). It has a baseline $\beta$ $($equation (\ref{eq13})$)$ as a mean of the rewards, which differs from the rewards based on the greedy decoding used in the earlier methods \cite{A3,A4}. The final policy gradient \cite{A20} to compute the reward for each step is calculated as follows:
\begin{equation}\label{eq12}
  \nabla_{\phi}L(\phi)= \\ -\frac{1}{k}\sum_{j=1}^{k}\left((r(S^j)-\beta)\nabla_{\phi} log(p_{\phi}(S ^j)\right) ,
\end{equation}
\begin{equation}\label{eq13}
\hspace*{-4.2cm}\beta = \left(\sum_{j} r(S^j)\right)/k
\end{equation}
where $\phi$ is the learning parameters, $S_j$ is the $j^{th}$ sentence in the beam, $k$ is the beam size, and $p_{\phi}$ represents a policy: an ``action'' of predicting the next word.

The proposed architecture, LATGeO, is trained using captions of a specific length, $C$, which are represented as a vector using an embedding layer with dimensions of $d_{model}$ and fed into a decoder layer. Moreover, elements' order in the sequence is represented using positional encoding, added to decoder input. Positional encoding can be seen as a vector representation of numbers in the range $[1, C]$ with $d_{model}$ dimensional decoder. The model takes the output of previously generated words as input to generate the next word during the prediction phase. 

\section{Implementation and Evaluation} \label{impleval} 
\begin{table*}[htbp]
  \centering
  \caption{LATGeO evaluation using BLEU-1, BLEU-4, METEOR, ROUGE-L, and CIDEr-D scores on Karpathy's split MSCOCO test dataset $($all values are in percentage $(\%)$ $)$. $\bigoplus$ represents an ensemble model, and the rest are single models. Bold figures depict the best results. $^*$ represents values after training the model with the same data pre-processing as ours and with the provided code. $^\#$ utilizes Resnet-152 based visual features. } 
    \begin{tabular}{rl|l|l|l|l|ll}
    \cline{2-7}
       &\textbf{Model}  & \textbf{BLEU-1}  & \textbf{BLEU-4} & \textbf{METEOR} & \textbf{ROUGE-L} & \textbf{CIDEr-D} \\
    \cline{2-7}
    \cline{2-7}
    
  \hspace{2.5em}\ldelim\{{3}{*}[DNN-Based Models]   &NICv2$^{\bigoplus}$ \cite{A11} & -  & 32.1 & 25.7 & - & 99.8  \\
     &MSM $^\#$ \cite{A14} & 73.0 & 32.5 & 25.1 & - & 98.6 \\
    & LSTM\_p + ATT\_s \cite{A36} & 73.8  & 32.7 & 26.1 & 54.1 & 101.8\\
   
    \cline{2-7}
   \hspace{2.5em}\ldelim\{{1}{*}[Template-Based Model] & NBT \cite{A18} & 75.5  & 34.7 & 27.1 & - & 107.2 \\
    \cline{2-7}
     
 \hspace{2.5em}\ldelim\{{16}{*}[Attention-Based Models]   &ATT-FCN $^{\bigoplus}$ \cite{A53} & 70.9 &  30.4 & 24.3 & - & - \\
  
    &Hard-Attention \cite{A51} & 71.8 &  25.0 & 23.0 & - & -\\
     &Fine-Grain \cite{A55} & 71.2 &  26.5 & 24.7 & - & 88.2\\
    &SCA-CNN \cite{A52} & 71.9 &  31.1 & 25.0 & 53.1 & 95.2\\
     &Obj-R + Rel-A \cite{A46} & 73.2 & 32.8 & 25.6 & 53.4 & 96.5 \\
   & Bawg-LSTM+mean \cite{A54} & 71.9 &  30.2 & 25.3 & - & 99.8\\
   
   
    &GHA \cite{A44} & 73.3 & 32.1 & 25.5 & 53.8 & 99.9 \\
    &Up-Down\cite{A4} & 74.5 & 33.4 & 26.1 & 54.4 & 105.4\\
   
    
   &SCST $($Att2in$)$ $^{\bigoplus}$ \cite{A3} & - & 32.8 & 26.7 & 55.1 & 106.5\\
   & AttM\cite{A33} & 75.7 & 33.7 & 26.3 & 55.1 & 106.8\\
   & Adaptive $^{\bigoplus}$  \cite{A12} & 74.2  & 33.2 & 26.6 & - & 108.5\\

  & BiGr\_{rg} \cite{A57} & 76.2 & 35.0 & 27.0 & - & -\\  
  &  Stack-Cap $($C2F$)$\cite{A47} & 76.2 & 35.2 & 26.5 & - & 109.1\\
    
   & GateCap\_A\cite{A56}& 75.9 & 35.5 & 27.4 & 56.3 & 110.8\\
    & ARL\cite{A34}& 75.9 & 35.8 & \textbf{27.8} & 56.4 & 111.3\\
    &att-ref\cite{A45}& 76.4 & 36.1 & 27.6 & 56.4 & 114.5\\
   
      \cline{2-7}
\hspace{2.5em}\ldelim\{{3}{*}[Transformer-Based Models]  & Up-Down $+$ ObjRel-Trans\cite{A2} & 75.6 & 33.5 & 27.6 & 56.0 & 112.6  \\
   & MeshTrans$^*$\cite{A20} & 75.7 & 35.4 & 27.8 & 56.4 & 113.1\\

   \cline{2-7}
    &\textit{LATGeO $($Ours$)$} & \textbf{76.5} & \textbf{36.4} & \textbf{27.8} & \textbf{56.7} & \textbf{115.8}\\
    \cline{2-7}
    \end{tabular}%
    
  \label{withoutRL}%
\end{table*}%

\subsection{Dataset}
In this study, we use the MSCOCO dataset \cite{A16}. The dataset consists of $123,287$ labeled images, randomly split into $113,287$ images in the train set, $5,000$ each in the validation set, and the test set using the standard Karpathy split technique \cite{A17}. Each image in the dataset has $5$ different captions as target captions. For online testing, the split of the dataset is different, and there are $82,783$ images in training, $40,504$ images in the validation set, and $40,775$ images in the test set. Target captions of images for online testing are not available publicly.  The target captions are converted into lower-case, and each caption is limited to a length of $22$ words. For LAM, a dictionary is made of words that occurred more than five times in the whole corpus, resulting in a vocabulary size of $10,021$ distinct words. Less frequent words are substituted with the ``UNK'' keyword, and every sentence starts with ``START'' and ends with ``END'' keywords. 

\subsection{Implementation} \label{impl}
The proposals are generated with Faster R-CNN \cite{A22} with a base ResNet$-101$ \cite{A23}. Faster R-CNN is fine-tuned on the Visual Genome dataset \cite{A61,A20,A4}, which contains $1600$ object classes. In addition to the objects classes, this dataset provides annotations for objects' attributes $($like colors, sizes, etc.$)$ and their relationships $($like below, under, on, in, etc.$)$. However, this study only uses annotations of objects classes; the other available annotations are left for further experimentation by extending the proposed framework. Objects with class probabilities greater than $0.7$ are selected as proposals, and a maximum of $50$ objects per image are selected, similar to previous work \cite{A20}. Similarly, a $2048-$dimensional features vector for less significant details is extracted for each image using ResNet$-50$ \cite{A23}. Words are embedded using linear projection of one-hot vector representations of $512-$dimensions, which is the same as the input dimensions of our proposed transformer model. Moreover, sinusoidal encoding is used for positional encoding of words in a target caption \cite{A20}. In LATGeO, input and output dimensions of encoder-decoder architecture are set to $(d_{model}=512)$, the number of heads is set to $(h=8)$ in multi-head attention, and memory size $M=40$ in the encoder layer. The number of stacked encoders and decoders layers is $3$ $(L=3)$ (Consider \ref{AddE2}). 

LATGeO is trained on a machine with Nvidia $1080$Ti and RAM $16$GB, using masked cross-entropy objective function, XE $($equation (\ref{eq11})$)$, with label smoothing of $0.1$. We employ Adam optimizer with a learning rate scheduling strategy used in the vanilla transformer \cite{A1} with $10,000$ warmup iterations. After supervised learning in RL, the reward function in equation \ref{eq12} is used with the patience of $5$ based on the CIDEr-D score on the validation set from the Karpathy split technique. The reward is achieved by decoding sentences using beam search with a beam-size of $k=5$ and a learning rate of $5 * 10^{-6}$. All experiments are performed with a batch size of $50$, and early-stopping based on the CIDEr-D score is used for regularization. 
\begin{table*}[htbp]
  \centering
  \caption{LATGeO results after training with reinforcement learning $($CIDEr optimization$)$ $($all values are in percentage $(\%)$ $)$. $\bigoplus$  an ensemble model, and the rest are single models. $^*$ represents values after training the model with the same data pre-processing as ours and with the provided code. $^\#$ utilizes Resnet-101 based visual features.} 
    \begin{tabular}{rl|l|l|l|l|l|l}
   \cline{2-8}
       &\textbf{Model}  & \textbf{BLEU-1}  & \textbf{BLEU-4} & \textbf{METEOR} & \textbf{ROUGE-L} & \textbf{SPICE} & \textbf{CIDEr-D} \\
     \cline{2-8}
     \cline{2-8}
    \hspace{2.5em}\ldelim\{{4}{*}[DNN-Based Models        ]& 
    RL-EmbeddedReward \cite{A58} & 71.3 & 30.4 & 25.1 & 52.5 &   - & 93.7\\
     &RL-G-GAN\cite{A59} & - & 29.9 & 24.8 & 52.7 & 19.9  & 102.0\\
      
   & Actor-Critic  \cite{A13}&-& 34.4 & 26.7 & 55.8 &-& 116.2\\

   & SGAE$^{\bigoplus}$ \cite{A8} & \textbf{81.0} & \textbf{39.0} & 28.4 & \textbf{58.9} & 22.2  & 129.1\\
    
    
   
    \cline{2-8}
  \hspace{2.5em}\ldelim\{{12}{*}[Attention-Based Models]   &Hierarchical-Attention \cite{A49} & 73.0 & 28.6 & 25.3 & 56.5 & -  &  92.5\\

     &SCST $($Att2all$)$ $^{\bigoplus}$  \cite{A3} & - & 35.4 & 27.1 & 56.6 & -  & 117.5\\
    &Up-Down\cite{A4} & 79.8 & 36.3 & 27.7 & 56.9 &  21.4 & 120.1\\
    &Obj-R + Rel-A\cite{A46}& 79.2 & 36.3 & 27.6 & 56.8 &  21.4 & 120.2\\
    &Stack-Cap $($C2F$)$\cite{A47} & 78.6 & 36.1 & 27.4 & 56.9 & 20.9  & 120.4\\
    &GateCap\_O\cite{A56}& 79.3 & 37.3 & 27.9 & 57.7 & -  &  124.0\\
    &hLSTMat$^\#$ \cite{A60}& 79.9 & 37.5 & 28.5 & 58.2 &  22.3 & 125.6\\
    &RFNet$^{\bigoplus}$ \cite{A7}& 80.4 & 37.9 & 28.3 & 58.3 &  21.7 & 125.7\\
    & Fine-Visual-policy\cite{A30} & - & 38.6 & 28.3 & 58.5 & 21.6  & 126.3\\
    
 &Up-Down+HIP \cite{A9}& -& 38.2 &   28.4   & 58.3 &  -   & 127.2 \\
    &GCN-LSTM$^{\bigoplus}$ \cite{A6} & 80.9 & 38.3 & 28.6 & 58.5 & 22.1  & 128.7\\
    & SGAE -KD\cite{A40} & \textbf{81.0} & 38.8 & 28.8 & 58.8 & 22.4   & 129.6\\
   \cline{2-8}
  \hspace{2.5em}\ldelim\{{4}{*}[Transformer-Based Models]   &ObjRel-Trans\cite{A2} & 80.5 & 38.6 & 28.7 & 58.4 & 21.2  & 128.3\\
     &VRAtt-Soft-Trans \cite{A31}& 80.5 & 38.5 & 28.9 & \textbf{61.8} & 22.8  & 129.2\\
   & MeshTrans$^*$\cite{A20} & 80.7 & 38.8 & 28.9 & 58.4 & 22.6  & 129.2\\
    & AoANet\cite{A19} & 80.2 & 38.9 & \textbf{29.2} & 58.8 & 22.1   & 129.8\\
    \cline{2-8}
&\textit{LATGeO $($Ours$)$} & \textbf{81.0} & 38.8 & \textbf{29.2} & 58.7 &  \textbf{22.9}  & \textbf{131.7}\\
    \cline{2-8}
    \end{tabular}%
  \label{withRL}%
\end{table*}%
\begin{figure*}
\begin{center}
		\includegraphics[height=12.0cm, width=16.0cm]{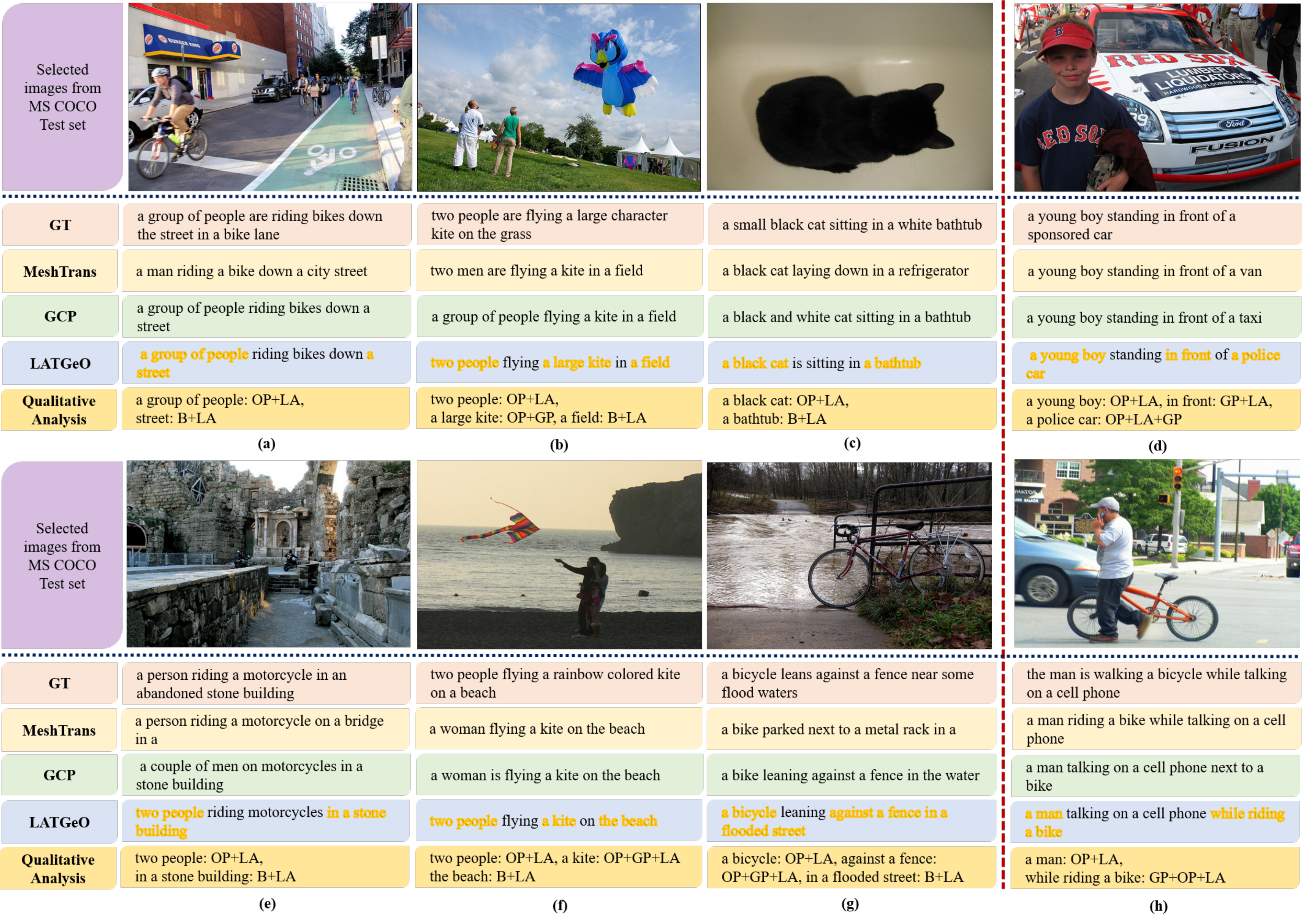}
\end{center}
\caption{Qualitative results of selected images from MSCOCO test dataset. GT, MeshTrans, and GCP represent captions generated by ground truth, MeshTrans \cite{A20}, and graphically coherent proposals module of LATGeO, respectively. Whereas OP, LA, B, and GP represent object proposals, label-attention, background, and geometrical proposals modules of LATGeO, respectively. $($a$) \textendash ($c$)$ and $($e$) \textendash ($g$)$ represent successful image captioning cases, and $($d$)$, $($h$)$ represent the case where LATGeO generated captions are not very accurate. Last rows (Qualitative Analysis) brief the expected modules of the LATGeO engaged for effective caption generation and show the improvement compared to the MeshTrans model. Highlighted words in the LATGeO captions represent caption refinements.}
\label{quality}
\end{figure*}
\begin{figure}
\begin{center}
		\includegraphics[trim={7.5cm 7.7cm 11.0cm 2.0cm}, clip=true, height=5.0cm, width=7.8cm]{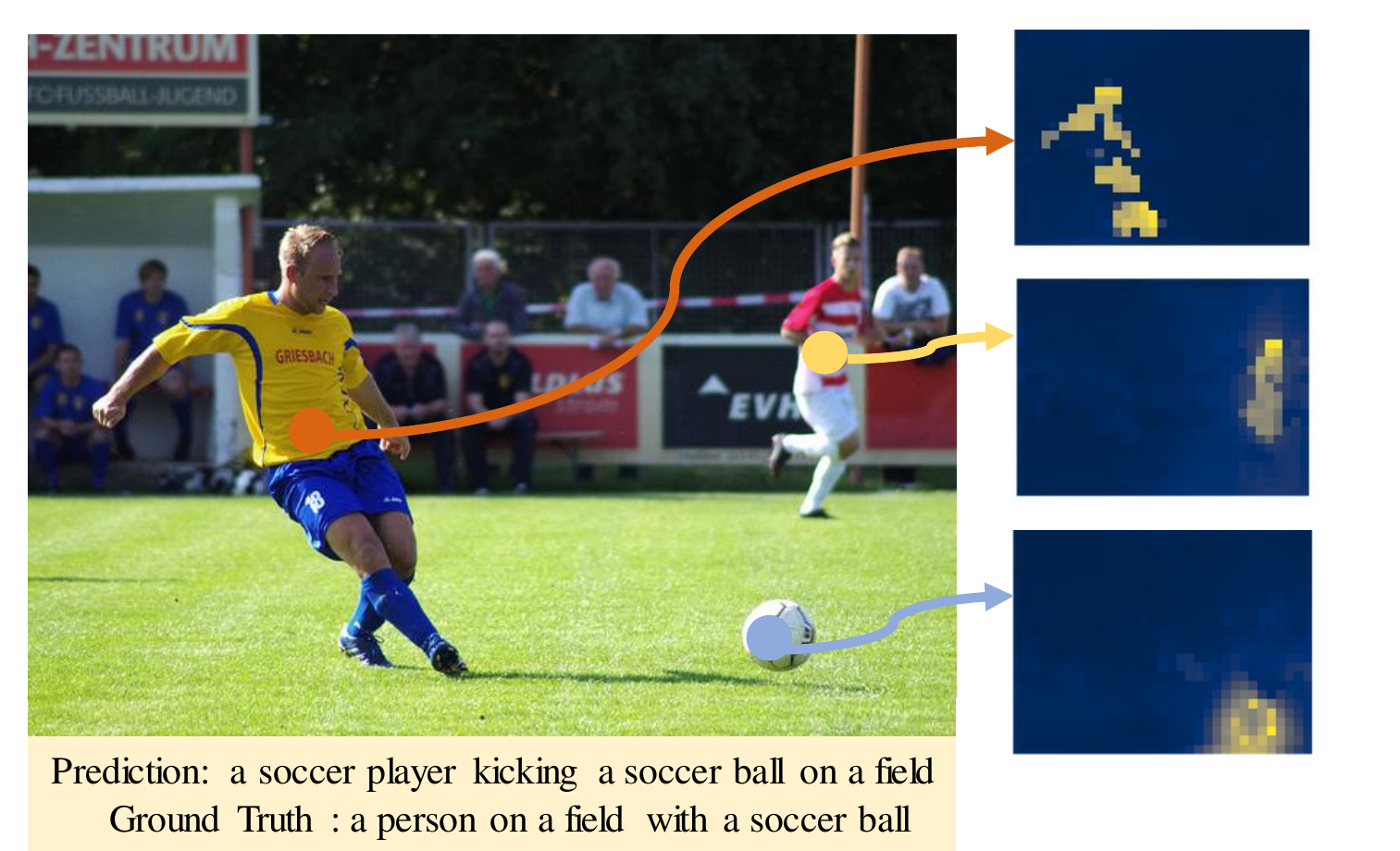}
	\end{center}
\caption{Demonstration of the self-attention module of LATGeO, where the arrows with orange, yellow, and blue colors exhibit the attention maps of the associated objects in the provided image.}
\label{sa}
\end{figure}
\begin{table*}[htbp]
  \centering
  \caption{A comparison of different modules proposed in LATGeO,  where \textcircled{\raisebox{-0.9pt}{1}} : Object Proposal Module--Faster RCNN,  
  \textcircled{\raisebox{-0.9pt}{2}} : Background Features,
  \textcircled{\raisebox{-0.9pt}{3}} : Geometrically Coherent Proposals: Objects shapes' L1-Comparison, \textcircled{\raisebox{-0.9pt}{4}} : Geometrically Coherent Proposals: Objects shapes' Ratio-Comparison,
  \textcircled{\raisebox{-0.9pt}{5}} : Label-Attention,
  \textcircled{\raisebox{-0.9pt}{6}} : Fully-Connected Transformer,
  \textcircled{\raisebox{-0.9pt}{7}} : Cross-Entropy Loss + RL. Bold figures stand for the best performance in all. } 
    \begin{tabular}{c|c|c|c|c|c|c}
    \hline
       \textbf{Model}  & \textbf{BLEU-1}  & \textbf{BLEU-4} & \textbf{METEOR} & \textbf{ROUGE-L} & \textbf{CIDEr-D}  & \textbf{SPICE} \\
    \hline
    \hline
    \textcircled{\raisebox{-0.9pt}{1}} + \textcircled{\raisebox{-0.9pt}{6}} +
        \textcircled{\raisebox{-0.9pt}{7}} & 80.5 & 38.6 & 28.7 & 58.4 & 129.2 & 22.5\\
      \textcircled{\raisebox{-0.9pt}{1}} + \textcircled{\raisebox{-0.9pt}{3}} + \textcircled{\raisebox{-0.9pt}{6}} + \textcircled{\raisebox{-0.9pt}{7}} & 80.3 & 38.4 & 28.9 & 58.6 & 129.9 & 22.7\\
       \textcircled{\raisebox{-0.9pt}{1}} + \textcircled{\raisebox{-0.9pt}{4}} +
        \textcircled{\raisebox{-0.9pt}{6}} +\textcircled{\raisebox{-0.9pt}{7}}  & 80.5 & 38.5 & 29.2 & 58.4 & 130.4 & 22.7\\
        \textcircled{\raisebox{-0.9pt}{1}} + \textcircled{\raisebox{-0.9pt}{2}} +
        \textcircled{\raisebox{-0.9pt}{4}} +\textcircled{\raisebox{-0.9pt}{6}} + \textcircled{\raisebox{-0.9pt}{7}} & 80.6 & 38.9 & 29.2 & 58.5 & 130.7 & 22.9\\
             \textcircled{\raisebox{-0.9pt}{1}} + \textcircled{\raisebox{-0.9pt}{2}} +
        \textcircled{\raisebox{-0.9pt}{4}} + \textcircled{\raisebox{-0.9pt}{5}} +
        \textcircled{\raisebox{-0.9pt}{6}} +
        \textcircled{\raisebox{-0.9pt}{7}} \textit{$($LATGeO$)$ $($Ours$)$} & \textbf{81.0} & \textbf{38.8} & \textbf{29.2} & \textbf{58.7} & \textbf{131.7} & \textbf{22.9}\\

    
    
  
    \hline
    \end{tabular}%
  \label{ablationStudy}%
\end{table*}%
\begin{table*}[htbp]
\centering
\caption{LATGeO results using Faster-RCNN and DETR}
\begin{tabular}{l|l|l|l|l|l|l|l|l|l|l|l|l}
\hline
\multicolumn{1}{c|}{\multirow{2}{*}{Model}} & \multirow{2}{*}{B-1} & \multirow{2}{*}{B-4} & \multirow{2}{*}{M} & \multirow{2}{*}{R} & \multirow{2}{*}{C} & \multicolumn{7}{c}{SPICE}                                  \\ \cline{7-13} 

\multicolumn{1}{c|}{}                       &                         &                         &                         &                          &                          & All & Object & Att & Relation & Color & Count & Size \\ \hline   \hline
    LATGeO-DETR (XE)                                       & 75.3                        &   34.7                      &     27.1                    &      55.7                    &  112.1                        & 20.3    &     37.1   &  9.8          &    5.6      &  10.0     &   11.8    & 4.7     \\ 
   LATGeO-Faster R-CNN  (XE)  &       76.5             &    36.4                    &         27.8                &    56.7                      &     115.8                     &  20.9  &   37.6    &   11.0         &   5.8       & 12.5     & 13.0      & 5.1     \\ \hline
   LATGeO-DETR  (RL)                                      & 79.8                        &   37.2                      &     28.5                    &      57.6                    &  127.0                        &  22.0   &   39.7     &  10.8          &   6.6       &    12.1   &   22.1    &   3.1   \\ 
   LATGeO-Faster R-CNN (RL) &      81.0             &    38.8                     &         29.2                &    58.7                      &     131.7                     &  22.9   &    40.7    &   12.1         &   7.1       & 14.6      & 22.9      & 4.2     \\ \hline
\end{tabular}
\label{detrrcnn}%
\end{table*}

\begin{table*}[htbp]
\centering
  \caption{Online evaluation of LATGeO on MSCOCO test server. $^*$ utilizes Resnet-152 based visual features. $^\#$ utilizes RestNet-101 based visual features. The results are sorted on the CIDEr-D  values. Only single model architectures are reported in this table. } 
 \begin{tabular}{lllllllllllllll} 
\hline
\textbf{~ ~ ~ ~ Model}                                         & \multicolumn{2}{l}{\textbf{BLEU-1}} & \multicolumn{2}{l}{\textbf{BLEU-2}} & \multicolumn{2}{l}{\textbf{BLEU-3}} & \multicolumn{2}{l}{\textbf{BLEU-4}} & \multicolumn{2}{l}{\textbf{METEOR}} & \multicolumn{2}{l}{\textbf{ROUGE-L }} & \multicolumn{2}{l}{\textbf{CIDEr-D}}  \\ 
\hline
                                                               & c5     & c40                        & c5     & c40                        & c5     & c40                        & c5      & c40                       & c5     & c40                        & c5     & c40                          & c5      & c40                         \\ 
\hline
\hline
Hard-Attention \cite{A51}~                                 & 70.5~  & 88.1~                       & 52.8~ & 77.9~                      & 38.3~  & 65.8~                      & 27.7~   & 53.7~                     & 24.1~  & 32.2~                       & 51.6~ & 65.4~                        & 86.5~   & 89.3~                        \\
GHA\cite{A44}~                                                  & 72.9~ & 93.7~                     & 56.0~ & 81.8~                     & 41.9~ & 70.8~                     & 31.3~  & 59.8~                     &25.2~ & 34.1~                     & 53.3~ & 68.3~                      & 95.4~  & 96.3~                      \\
AttM\cite{A33}~                                                  & 75.5~  & 92.4~                      & 58.8~  & 84.6~                      & 44.5~  & 74.2~                       & 33.3~  & 62.9~                     & 26.0~  & 34.7~                      & 54.8~   & 69.7~                       & 103.1~  & 104.7~                       \\
ARL \cite{A34}~                                                   & -~      & -~                          & 58.9~   & 85.6~                       & 45.0~   & 75.6~                       & 34.3~    & 64.7~                      & 27.0~   & 36.4~                       & 55.5~   & 71.0~                         & 106.1~   & 106.4~                      \\
Actor-Critic (single)\cite{A13}~                                         & 77.8~ & 92.9~                     & 61.2~ & 85.5~                     & 45.9~ & 74.5~                      & 33.7~ & 62.5~                    & 26.4~ & 34.4~                     &55.4~ &69.1~                       & 110.2~  & 112.1~                       \\

Obj-R + Rel-A \cite{A46}~                                       & 79.2~   & 94.4~                       & 62.6~   & 87.2~                       & 47.5~   & 77.1~                       & 35.4~    & 65.8~                      & 27.3~   & 36.1~                       & 56.2~   & 71.2~                         & 115.1~   & 117.3~                       \\
MSM$^*$ \cite{A14}~   & 78.7~  & 93.7~                      & 62.7~  & 86.7~                      & 47.6~  & 76.5~                      & 35.6~   & 65.2~                     & 27.0~    & 35.4~                      & 56.4~   & 70.5~                       & 116.0~    & 118.0~                         \\


Stack-Cap (C2F)\cite{A47}~                                   & 77.8~  & 93.2~                       & 61.6~ & 86.1~                      & 46.8~  & 76.0~                      & 34.9~    & 64.6~                    & 27.0~  & 35.6~                       & 56.2~ & 70.6~                        & 114.8~  & 118.3~                       \\

hLSTMat$^\#$ \cite{A60}~                  & 79.4~  & 94.4~                      & 63.5~   & 88.0~                      & 48.7~  & 78.4~                     & 36.8~    & 67.4~                    & 28.2~   & 37.0~                      & 57.7~  & 72.2~                       & 120.5~  & 122.0~                       \\ 
Fine-Visual-policy \cite{A30}~                                   & 80.1~  & 94.9~                      & 64.7~  & 88.8~                      & 50.0~  & 79.7~                      & 37.9~   & 69.0~                     & 28.1~  & 37.0~                      & 58.2~  & 73.1~                         & 121.6~  & 123.8~                     \\

GCN-LSTM\cite{A6}~                                              & -~      & -~                          & \textbf{65.5}~  & 89.3~                      & \textbf{50.8}~  & 80.3~                      & \textbf{38.7}~   & 69.7~                     & 28.5~  & 37.6~                      & \textbf{58.5}~  & 73.4                         & 125.3~ & 126.5                       \\
SGAE -KD\cite{A40}~                                             & - ~     & -~                          & -~      & -~                          & 50.1~  & 79.9~                       & 38.2~   & 69.3~                    & 28.7~   & 37.9~                     & 58.4~   & \textbf{73.5}~                        & 124.5~ & 126.6~                       \\


\hline
\textit{LATGeO(Ours)}~                               & \textbf{80.5}~   & \textbf{95.4}~                      & 64.8~   & \textbf{89.6}~                       & 50.0~   & \textbf{80.8}~                       & 37.9~    & \textbf{70.3}~                      & \textbf{28.8} ~  & \textbf{38.2}~                       & 58.1~   & 73.2~                         & \textbf{126.7}~   & \textbf{130.1} ~                      \\
\hline
\end{tabular}
  \label{onlineeval}%
\end{table*}%
\begin{figure*}
\begin{center}
		\includegraphics[height=9.5cm, width=15.0cm]{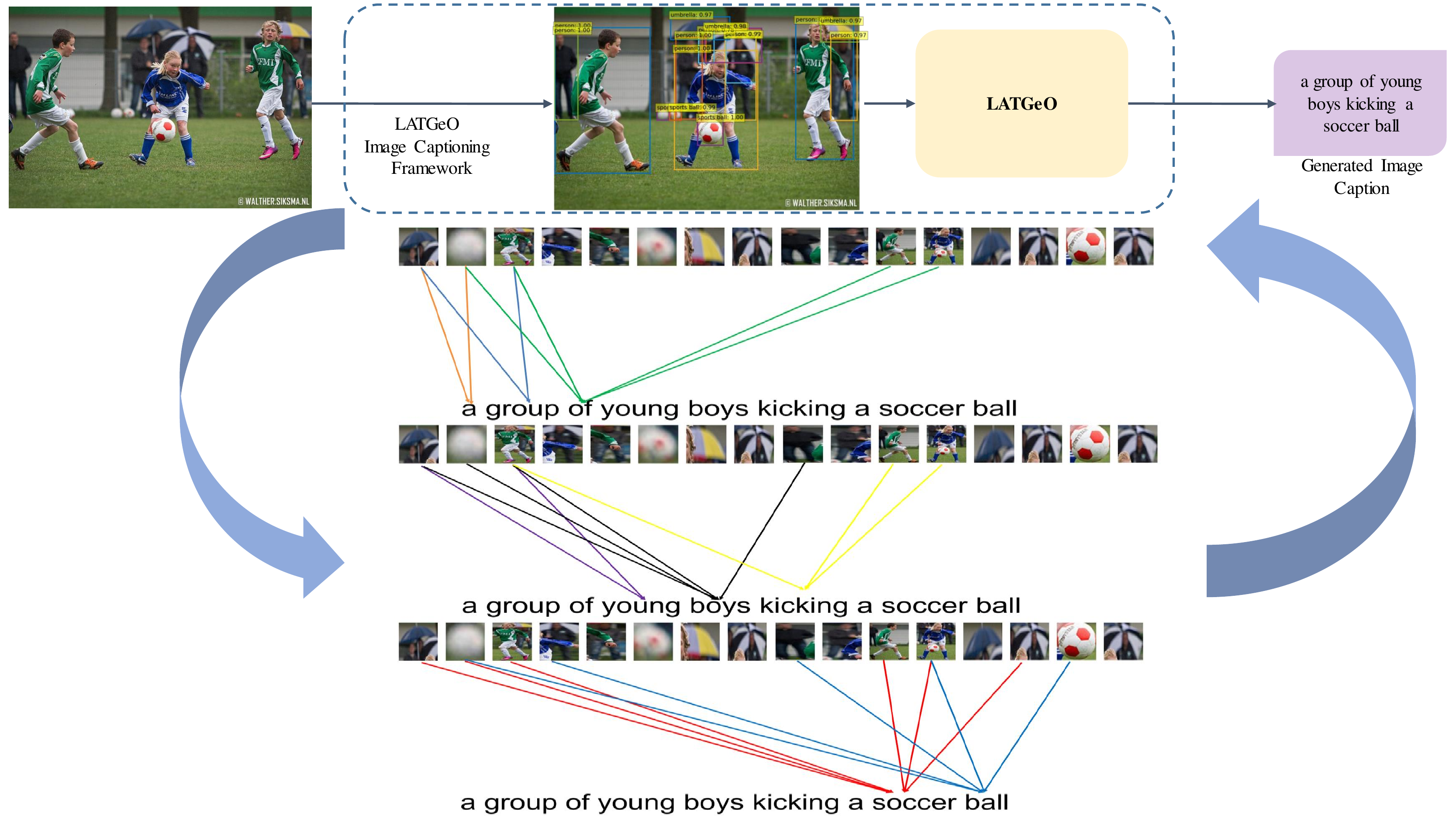}
	\end{center}
\caption{Demonstration of extracted objects' association to words using the proposed attention mechanism.}
\label{associate}
\end{figure*}
\subsection{Evaluation} \label{eval}

\subsubsection{Evaluation Metrics} \label{em}
We have evaluated the proposed architecture's performance using regularly used evaluation metrics, i.e., BLEU-1, BLEU-4 \cite{A62}, METEOR \cite{A63}, ROUGE-L \cite{A64}, SPICE \cite{A65}, and CIDEr-D \cite{A66}. The qualitative results on the MSCOCO $2014$ test set from the Karapathy split and the MSCOCO server evaluation test set are given in Table~\ref{withoutRL}, Table~\ref{withRL}, and Table~\ref{onlineeval}.  

\subsubsection{Evaluation on MSCOCO Karapathy Split} \label{oe}
LATGeO is compared with the recent best single-model algorithms as well as with recent ensemble-model algorithms. The proposed model trained using XE objective function given in equation (\ref{eq11}) outperforms all SOTA single-model and ensemble-model algorithms, as shown in Table~\ref{withoutRL}. Furthermore, it improves scores for all evaluation metrics compared to the transformer-based algorithms such as MeshTrans \cite{A20} and ObjRel-Trans \cite{A2}, i.e., $2.7\%$ and $3.2\%$ improvement on CIDEr-D scores, respectively. For a fair comparison, we have trained MeshTrans \cite{A20} with a similar preprocessing and hyper-parameters to our training of LATGeO. 

Furthermore, LATGeO, when trained with RL, boosts the performance and produces the highest BLEU-1, METEOR, SPICE, and CIDEr-D scores, as presented in Table~\ref{withRL}. LATGeO outperforms DNN, attention-based models, and transformer-based algorithms \cite{A2,A20,A19,A31} in most of the evaluation metrics. As shown, LATGeO outperforms the MeshTrans \cite{A20} in all evaluation metrics with a $2.5\%$ CIDEr score improvement and shows the superiority of our model over MeshTrans. It also outperforms another transformer-based algorithm \cite{A2}, which includes geometrical features different from ours with $3.4\%$ CIDEr score improvement and $2.0\%$ CIDEr score improvement compared to \cite{A19}. Additionally, LATGeO outperforms transformer-based algorithm \cite{A31} in all metrics except ROUGE-L, with a $2.5\%$ CIDEr score improvement. Moreover, \cite{A8} gives a better METEOR score than LATGeO, and the possible reasons could be that \cite{A8} uses an ensemble technique to present their evaluation metrics. However, our proposed LATGeO single-model shows better evaluation results than \cite{A8} on other metrics, including CIDEr-D, where we achieve $2.6\%$ improvement.

\subsection{Discussion} \label{abs}

\subsubsection{DETR Objects Proposals} \label{DETR}
We have also experimented LATGeO with proposals generated using DETR \cite{A15} as it has several advantages over Faster-RCNN. Object proposal is one of the essential parts of our proposed technique. In this study, ResNet-$50$ \cite{A23} is used as a base network for DETR. The object proposals are passed through another ResNet-$50$ model to generate $2048-$dimensional visual feature maps.\\

Table~\ref{detrrcnn} presents a performance comparison between DETR and Faster R-CNN object detectors with LATGeO using XE objective function and RL. LATGeO with Faster R-CNN outperforms LATGeO with DETR because Faster R-CNN was fine-tuned on the Visual Genome dataset \cite{A61}, which connects the visual domain to language domain with $1600$ classes of objects, whereas DETR has $91$ object classes. A fine-tuned DETR model on the Visual Genome dataset for the image captioning may achieve better results than Faster R-CNN and left for future exploration. 

\begin{table}[htbp]
  \centering
  \caption{LATGeO evaluation with SPICE. It shows significant improvement in Relation, Attributes, Object, and Count metrics.} 
    \begin{tabular}{l|l|l|l|l|l|l|l}
    \hline
       \multicolumn{1}{c|}{\multirow{2}{*}{Model}} & \multicolumn{7}{c}{SPICE} \\ \cline{2-8} 
\multicolumn{1}{c|}{}  & \textbf{All}  & \textbf{Obj}  & \textbf{Att}  & \textbf{Rel}  & \textbf{Color} &  \textbf{Count} & \textbf{Size} \\

    \hline
    \hline
    Standard Transformer & 21.1 & 38.6 & 9.6 & 6.3  & 9.2 & 17.5 & 2.0 \\
    ObjRel-Trans \cite{A2} & 21.2 & 37.9 & 11.4 & 6.3  &  \textbf{15.5} & 17.5 &  \textbf{6.4}\\
    Up-Down \cite{A4} &21.4 & 39.1 & 10.0 & 6.5 & 11.4 & 18.4 & 3.2 \\
    hLSTMat \cite{A60} & 22.3& 40.3 & 11.2& 6.4& 15.2 & 14.4& 3.7\\
    MeshTrans \cite{A20}  & 22.6 & 40.0 & 11.6 & 6.9 & 12.9 & 20.4 & 3.5 \\
   \textit{LATGeO $($Ours$)$} & \textbf{22.9} & \textbf{40.7} & \textbf{12.1} & \textbf{7.1} &14.6 & \textbf{22.9} & 4.2 \\
    \hline
    \end{tabular}%
  \label{ablationStudy2}%
\end{table}%

\subsubsection{Ablation Study: Effectiveness of LATGeO Modules} \label{module}
In addition, Table~\ref{ablationStudy} illustrates the effectiveness of individual modules proposed in LATGeO. It asserts that modules \textcircled{\raisebox{-0.9pt}{1}}, \textcircled{\raisebox{-0.9pt}{3}}, \textcircled{\raisebox{-0.9pt}{4}}, \textcircled{\raisebox{-0.9pt}{6}}, and \textcircled{\raisebox{-0.9pt}{7}} collectively produce the highest scores. Moreover, Table~\ref{ablationStudy2} demonstrates the effectiveness of GCP module along with the LAM by decomposing the SPICE metric into objects, attributes, relation, color, count, and size metrics. We have compared these metrics with our recent transformer-based model, MeshTrans \cite{A20} and other recent methods. As shown, LATGeo shows improvements in relation, attribute, count, and object metrics, compared to all other mentioned methods, though outperforms in all metrics compared to the MeshTrans model. Tables~\ref{ablationStudy} and ~\ref{ablationStudy2} show that GCP and LAM improve the overall performance of LATGeO and generate fine captions.

\subsubsection{Qualitative Analysis of LATGeO} \label{quality}
We have shown image captions of selected images generated by the proposed framework in Fig.~\ref{quality} for qualitative analysis. It shows that the LATGeO generates semantically and syntactically correct sentences. Fig. \ref{quality} displays the results of LATGeO, and a qualitative analysis row is appended to relate the highlighted improvements of the captions. The qualitative analysis row speculates a most likely contribution in improvements from the proposed modules in this study. Fig.~\ref{sa} demonstrates the self-attention module of LATGeO, where each attention map of a corresponding object displays the object's significance compared to other objects in the tested image. The figure also shows the predicted caption and the ground truth caption. The significance of the attention mechanism is shown in Fig.~\ref{associate}, where we have demonstrated the association of extracted objects with particular words. Despite having many objects per image, LATGeO can adequately generalize to map only a small number of objects per word.

\subsubsection{Evaluation on Online COCO Server} \label{online}
Table~\ref{onlineeval} illustrates the online performance of our proposed architecture, LATGeO, on the COCO test server. We have utilized single-model LATGeO for the online evaluation. For a fair comparison, we have summarized the comparison of our model only with the top-performing single-models from the server leader-board. Moreover, as per our knowledge, our proposed framework, LATGeO, is the first to report a transformer-based single-model for online evaluation. Table~\ref{onlineeval} demonstrates that our model surpasses all the current state-of-the-art methods on most of the evaluation metrics and achieves an improvement of $3.5\%$ CIDEr score to the previous best single-model algorithms \cite{A40,A6}.

\section{Conclusions and Future Work} \label{con}
This study demonstrates an image captioning technique named LATGeO, which explores the utility of object identity preservation along with surrounding information to generate meaningful captions of still images. LATGeO binds objects' features, surroundings, geometrical properties, and associated labels of semantically coherent objects using a transformer. The proposed architecture generates proposals using Faster R-CNN and computes their geometrical coherence that helps the transformer to attend similar labels for a situation depicted in an image. The encoder and the associations of particular features of the objects are further reviewed, strengthen, and transcribed using labels of the detected objects by a decoder. The labels are a mapping of classes to the dictionary words using an attention layer. An extrinsic definition of proposals helped LATGeO to outperform SOTA algorithms on the MSCOCO test dataset. The proposed technique is trained with cross-entropy loss and fine-tuned with reward-based reinforcement learning, which improves the results and scores better than many SOTA offline ensembles and shows outstanding performance in the online evaluation. We explore objects detection for proposals, and other choices of proposals can be explored and explicitly guide the transformers. Similarly, object coherence and label generation can be further explored to improve the results. 

\section*{Acknowledgment}
This work was partially supported by the Institute of Information \& communications Technology Planning \& Evaluation (IITP) grant funded by the Korea government (MSIT) (No. 2014-3-00077, AI National Strategy Project) and by the Ministry of Culture, Sports, and Tourism (MCST) and Korea Creative Content Agency(KOCCA) in the culture Technology (CT) Research Development Program (R20200600020) 2021.


\appendices

\section{Additional Experimentation}
   \label{FirstAppendix}
\begin{table*}[htbp]
  \centering
  \caption{LATGeO evaluation with various compositions of connectivity.} 
 \begin{tabular}{ccllllll}
\hline
\multicolumn{2}{c|}{Model}                                                                                     & \multicolumn{1}{c}{\multirow{2}{*}{BLEU-1}} & \multicolumn{1}{l}{\multirow{2}{*}{BLEU-4}} & \multicolumn{1}{l}{\multirow{2}{*}{METEOR}} & \multicolumn{1}{l}{\multirow{2}{*}{ROUGE-L}} & \multirow{2}{*}{CIDEr-D} & \multirow{2}{*}{SPICE} \\ \cline{1-2}
Types of Layer-Connections                                                        & \multicolumn{1}{l|}{Layers} & \multicolumn{1}{c}{}                    & \multicolumn{1}{l}{}                    & \multicolumn{1}{l}{}                   & \multicolumn{1}{l}{}                   &                    &                    \\ \hline
\hline
Single-Connection                                                                 & \multicolumn{1}{c|}{\multirow{4}{*}{3}}       & 80.4                                     & 38.8                                     & \textbf{29.2}                                    & 58.5                                    & 129.5              &        22.9            \\ 
Skipped-Connection                                                                &                          \multicolumn{1}{c|}{}    & 80.0                                     & 38.3                                     &  29.1                                       & 58.3                                    & 128.8              &       \textbf{23.1}             \\
\begin{tabular}[c]{@{}c@{}}Residual-Connection \\ in Encoder\end{tabular}         &                         \multicolumn{1}{c|}{}     & 80.7                                     & \textbf{39.0}                                     &  29.0                                       & 58.5                                    & 128.4              &          22.8          \\ 
\begin{tabular}[c]{@{}c@{}}Residual-Connection\\ in Encoder Decoder\end{tabular}  &                         \multicolumn{1}{c|}{}     & \multicolumn{1}{l}{80.2}                & \multicolumn{1}{l}{38.8}                & \multicolumn{1}{l}{\textbf{29.2} }                  & \multicolumn{1}{l}{58.5}               & 129.3              &           \textbf{23.1}         \\ \hline 
\begin{tabular}[c]{@{}c@{}}Residual-Connection \\ in Encoder\end{tabular}         & \multicolumn{1}{c|}{\multirow{2}{*}{6}}   & 80.9                                     & 38.7                                     &  28.7                                       & 57.9                                    & 130.0              &         22.1           \\ 
\begin{tabular}[c]{@{}c@{}}Residual-Connection\\ in Encoder Decoder\end{tabular}  &                           \multicolumn{1}{c|}{}    & \multicolumn{1}{l}{80.6}                & \multicolumn{1}{l}{38.4}                & \multicolumn{1}{l}{29.0}                   & \multicolumn{1}{l}{58.2}               & 130.6              &            22.5        \\ \hline
Fully-Connected                                                           &                       \multicolumn{1}{c|}{6}        & 80.6                                     & 38.1                                     &   29.1                                      & 58.1                                    & 129.2              &            22.8        \\ \hline
\begin{tabular}[c]{@{}c@{}}Fully-Connected\\ \textit{LATGeO (Ours)}\end{tabular} & \multicolumn{1}{c|}{3}                            & \multicolumn{1}{l}{\textbf{81.0}}                & \multicolumn{1}{l}{38.8}               & \multicolumn{1}{l}{\textbf{29.2}}                  & \multicolumn{1}{l}{\textbf{58.7}}               &\textbf{131.7}            &      22.9              \\ \hline
\end{tabular}
  \label{connections}%
\end{table*}%
  \subsection{Composition of Encoder-Decoder layers of proposed Transformer}
 \label{AddE}
 We have performed additional experiments to  illustrate the effect of different types of connectivity between encoder-decoder layers. Fig.~\ref{3_layer_res} and Fig.~\ref{6_layer_res} demonstrate the details for connections in 3-layers and 6-layer architectures, respectively. Fig.~\ref{3_layer_res} (a) shows the single-connection when one encoder output is passed as an input to the corresponding decoder layer. (b) shows skip-connection, when randomly selected few encoder layers output is passed as input to the decoder layer after sigmoid gating. (c) residual-connection \cite{A23} among encoder layer: when a residual connection is included among encoder layers along with a fully-connected transformer. (d) residual-connection in encoder and decoder layers: a residual connection is included among encoder layers and decoder layers along with a fully-connected transformer.  Fig.~\ref{6_layer_res} (a) and (b) represent similar connections in 6-layers architecture.
 
 Table \ref{connections} demonstrates the comparative analysis of using different connectivity in LATGeO, and our model with fully-connected encoder-decoder layers outperforms other mentioned connection techniques.
 
 \subsection{Number of Encoder-Decoder Layers}
  \label{AddE2}
\begin{figure*}
\begin{center}
\includegraphics[trim={0.0cm 0.1cm 0.0cm 0.0cm}, clip=true, height=9cm, width=13.0cm]{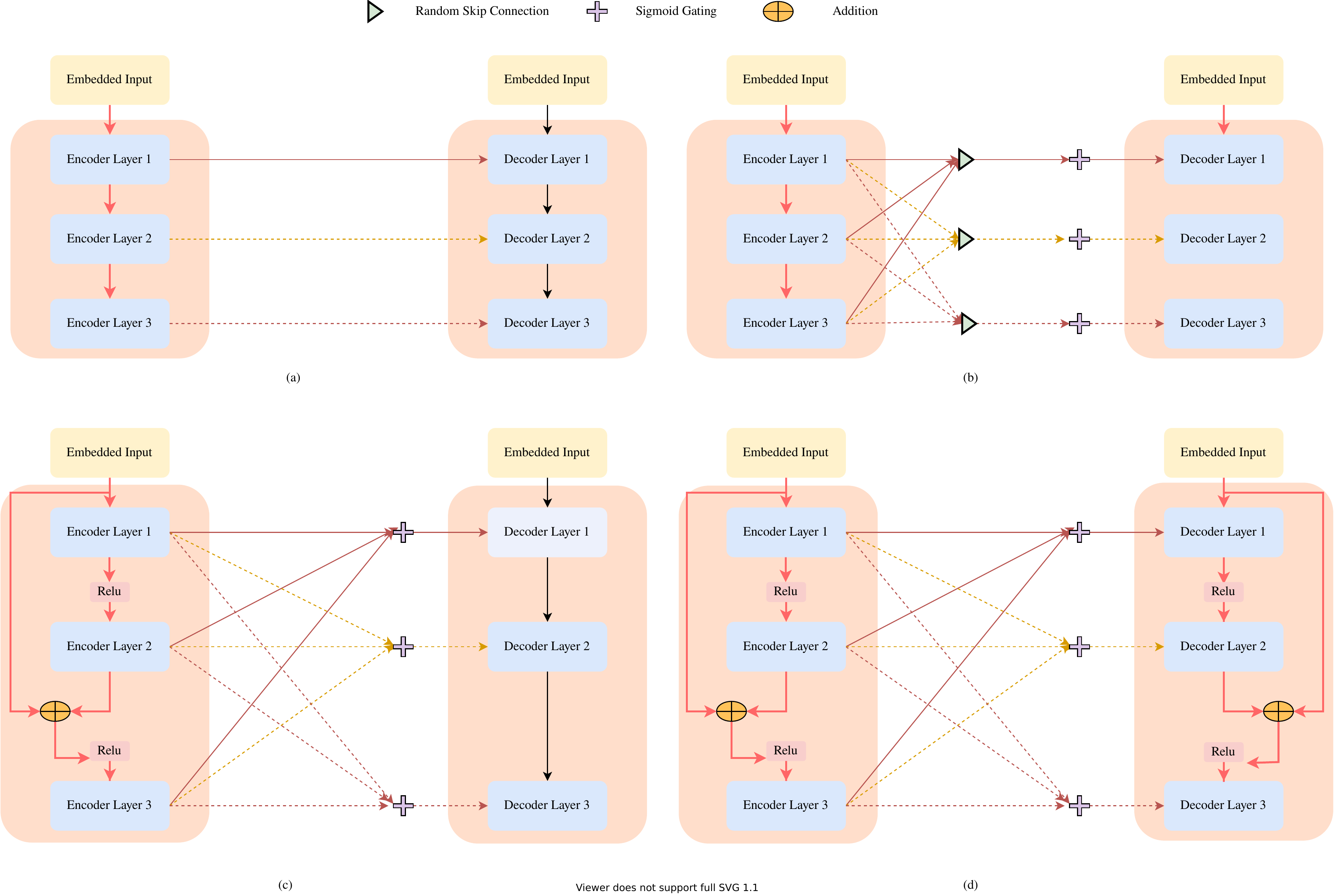}
	\end{center}
\caption{Compositions of connectivity between encoder and decoder of the transformer using 3-Layers. (a) Single-Connection. (b) Skip-Connection.(c) Residual-Connection among encoder layers with fully-connected layers. (d) Residual-connections in encoder and decoder layers with fully-connected layers.}
\label{3_layer_res}
\end{figure*}

\begin{figure*}
\begin{center}
\includegraphics[trim={0.0cm 0.1cm 0.0cm 0.0cm}, clip=true, height=9cm, width=15.0cm]{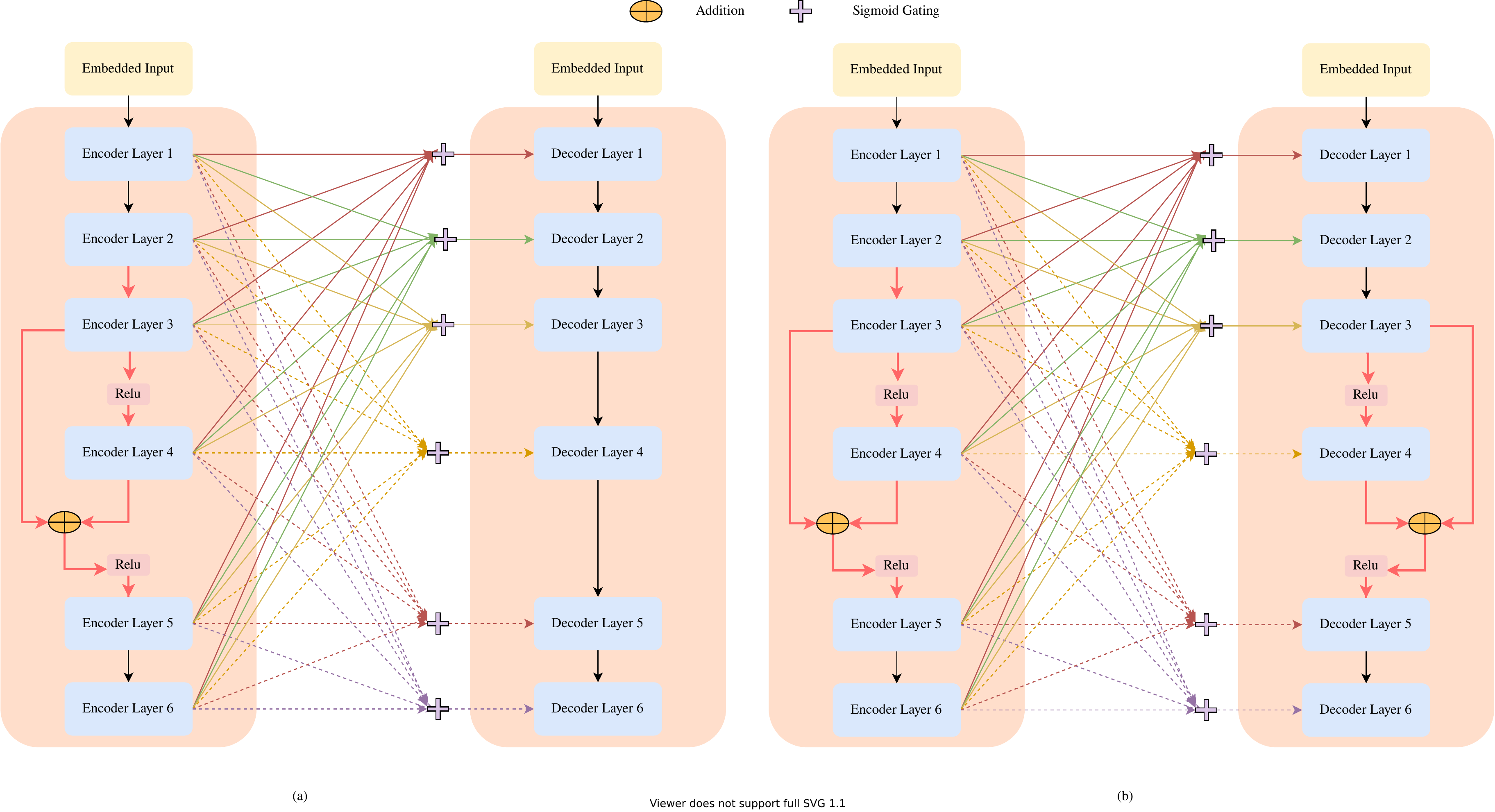}
	\end{center}
\caption{Compositions of connectivity between encoder and decoder of the transformer using 6-Layers. (a)
Residual-Connection among encoder layers along with fully-connected layers. (b) Residual-connection in encoder and decoder layers with fully-connected layers. }
\label{6_layer_res}
\end{figure*}

 Fig.~\ref{3_layer_res} and Fig.~\ref{6_layer_res} also demonstrate the proposed architecture with different numbers of encoder-decoder layers. Table \ref{connections} shows the effect of using 3-layers and 6-layers in the proposed algorithm, whereas using 3-layers of encoder-decoder and fully-connected transformer show the best results compared to other compositions.
  
\bibliographystyle{IEEEtranTIE}
\bibliography{arxiv.bib}

\end{document}